\documentclass[letterpaper, 10 pt, conference]{ieeeconf}  

\IEEEoverridecommandlockouts                              

\usepackage{graphics} 
\usepackage{epsfig} 
\usepackage{amsmath} 
\usepackage{amssymb}  
\usepackage[table]{xcolor}

\title{\LARGE \bf
Hoecken-D Hand: A Novel Robotic Hand for Linear Parallel Pinching and Self-Adaptive Grasping*
}

\author{Wentao Guo and Wenzeng Zhang, \textit{Member, IEEE}
\thanks{*Research supported by Foundation of \textit{Enhanced Student Research Training (E-SRT)} and Foundation of \textit{Open Research for Innovation Challenges (ORIC)}, X-Institute.}
\thanks{Wentao Guo is with Computer Science and Technology, Beijing Institute of Technology, China and Laboratory of Robotics, X-Institute, Shenzhen, China(email: yinsumirage@gmail.com).}
\thanks{Wenzeng Zhang is with Laboratory of Robotics, X-Institute, Shenzhen, China and Dept. of Mechanical Engineering, Tsinghua University, Beijing, China (Corresponding author, email: wenzeng75@163.com).}
}

\begin{document}

\maketitle
\thispagestyle{empty}
\pagestyle{empty}

\begin{abstract}

This paper presents the Hoecken-D Hand, an underactuated robotic gripper that combines a modified Hoecken linkage with a differential spring mechanism to achieve both linear parallel pinching and a mid-stroke transition to adaptive envelope. The original Hoecken linkage is reconfigured by replacing one member with differential links, preserving straight-line guidance while enabling contact-triggered reconfiguration without additional actuators. A double-parallelogram arrangement maintains fingertip parallelism during conventional pinching, whereas the differential mechanism allows one finger to wrap inward upon encountering an obstacle, improving stability on irregular or thin objects. The mechanism can be driven by a single linear actuator, minimizing complexity and cost; in our prototype, each finger is driven by its own linear actuator for simplicity. We perform kinematic modeling and force analysis to characterize grasp performance, including simulated grasping forces and spring-opening behavior under varying geometric parameters. The design was prototyped using PLA-based 3D printing, achieving a linear pinching span of approximately 200 mm. Preliminary tests demonstrate reliable grasping in both modes across a wide range of object geometries, highlighting the Hoecken-D Hand as a compact, adaptable, and cost-effective solution for manipulation in unstructured environments.

\end{abstract}

\section{INTRODUCTION}

Over the past few decades, robotic end-effectors have evolved from rigid industrial clamps to dexterous hands capable of adaptive manipulation in unstructured environments. Fully actuated dexterous hands, such as the Utah/MIT Hand's pneumatic tendon system~\cite{MIThand} and the modular DLR Hand~\cite{DLRhand2008}, offer remarkable manipulation versatility through independent joint control and sophisticated actuation schemes. However, their intrinsic complexity, substantial weight, and high manufacturing cost have limited their widespread deployment beyond specialized applications. Representative designs like the Stanford/JPL Hand~\cite{JPLhand1987} and NASA's Robonaut Hand~\cite{Robonauthand} have demonstrated high-performance manipulation and task adaptability, even in challenging domains such as space operations, but often at the expense of mechanical simplicity and ease of maintenance.

To address these challenges, underactuated mechanisms that trade active degrees-of-freedom (DOFs) for passive mechanical intelligence have been extensively explored. Early self-adaptive designs were classified into coupled, parallel and hybrid modes, with the parallel self-adaptive (PASA) mode receiving particular attention for its ability to achieve parallel fingertip motion~\cite{PASA2016}. The well-known PASA implementations include the Barrett Hand~\cite{Barrett}, the Robotiq adaptive gripper~\cite{robotiq}, and the SDM Hand~\cite{SDM}, each adopting distinct mechanical architectures to balance adaptability and robustness.

The PASA mode mitigates the limitations of traditional parallel grippers whose fingertip trajectories are inherently circular, preventing the grasp of thin-plate objects. To achieve near-linear fingertip motion, various linkages have been introduced, such as the Chebyshev linkage~\cite{chebyshev2017}, Hoecken linkage~\cite{Liu2019}, and rack–crank–slider mechanisms~\cite{Feng2024}. Recent work by Dukchan Yoon~\cite{Kim2022} integrated a Hart linkage with a parallelogram to achieve a fully passive, environment-adaptive robotic finger. Differential links, in particular, have been shown to enable passive transitions between grasp modes without sensing or control intervention~\cite{Liang2023}.

\begin{figure}[h]
    \centering
    \includegraphics[width=0.7\linewidth]{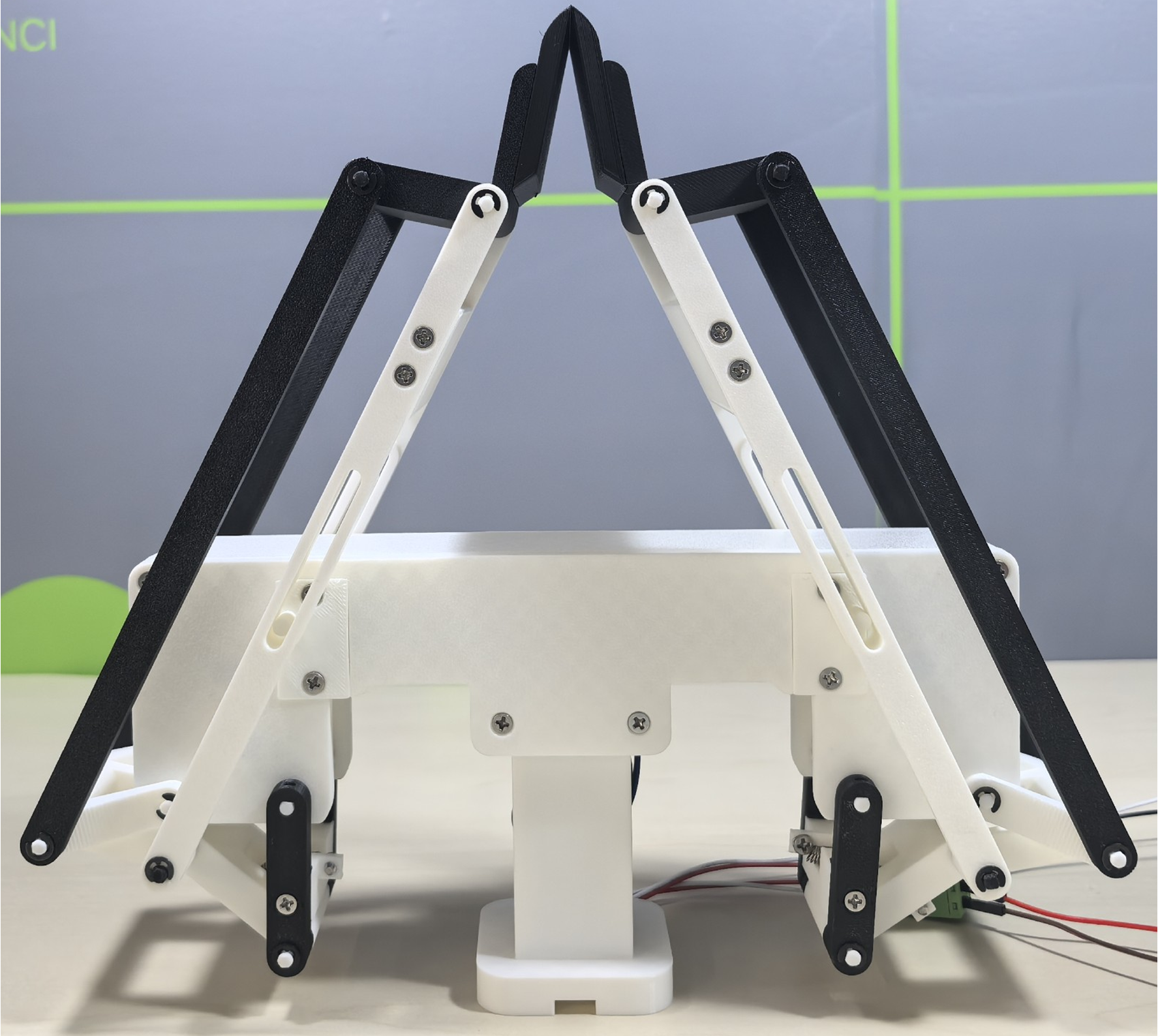}
    \caption{3D-printed prototype of the proposed Hoecken-D hand for linear pinching and self-adaptive grasping.}
    \label{fig:prototype}
\end{figure}

This paper introduces Hoecken-D Hand, a novel robotic finger design that combines a Hoecken linkage for near-straight-line fingertip motion with a differential linkage that passively enables an encompassing grasp when the finger contacts an object. In operation, the finger first performs linear parallel pinching for accurate grasping of thin or regular objects. When motion is blocked by a larger or irregular shape, the differential mechanism automatically reorients the fingertip into an adaptive enveloping posture. Recent work by Liu~\cite{Liu2023} also combined Hoecken and differential linkages but with limited wrapping range and a more complex multi-spring structure. Our design simplifies the mechanism and achieves a larger bottom-up rotation to improve vertical support and grasp stability. The mechanism can be driven by a single actuator for self-adaptive grasping of two fingers, but in our prototype each finger uses an independent actuator to keep the design simple and compact. A 3D-printed prototype (Fig.~\ref{fig:prototype}) was built, and both simulation and experiments confirm its robust grasping performance across objects of different sizes and shapes.

\section{DESIGN}

In this work, the proposed finger mechanism integrates a Hoecken linkage for near-linear horizontal motion with a differential linkage that passively enables the distal phalanx to transition into an enveloping configuration upon contact. This combination allows the finger to perform initial parallel pinching and, when necessary, adapt its orientation to conform to the object.

\subsection{Fingertip Translation: Hoecken Linkage for Linear Motion and Horizontal Compliance}

\begin {figure}[h]
    \centering
    \includegraphics [width=0.68\linewidth]{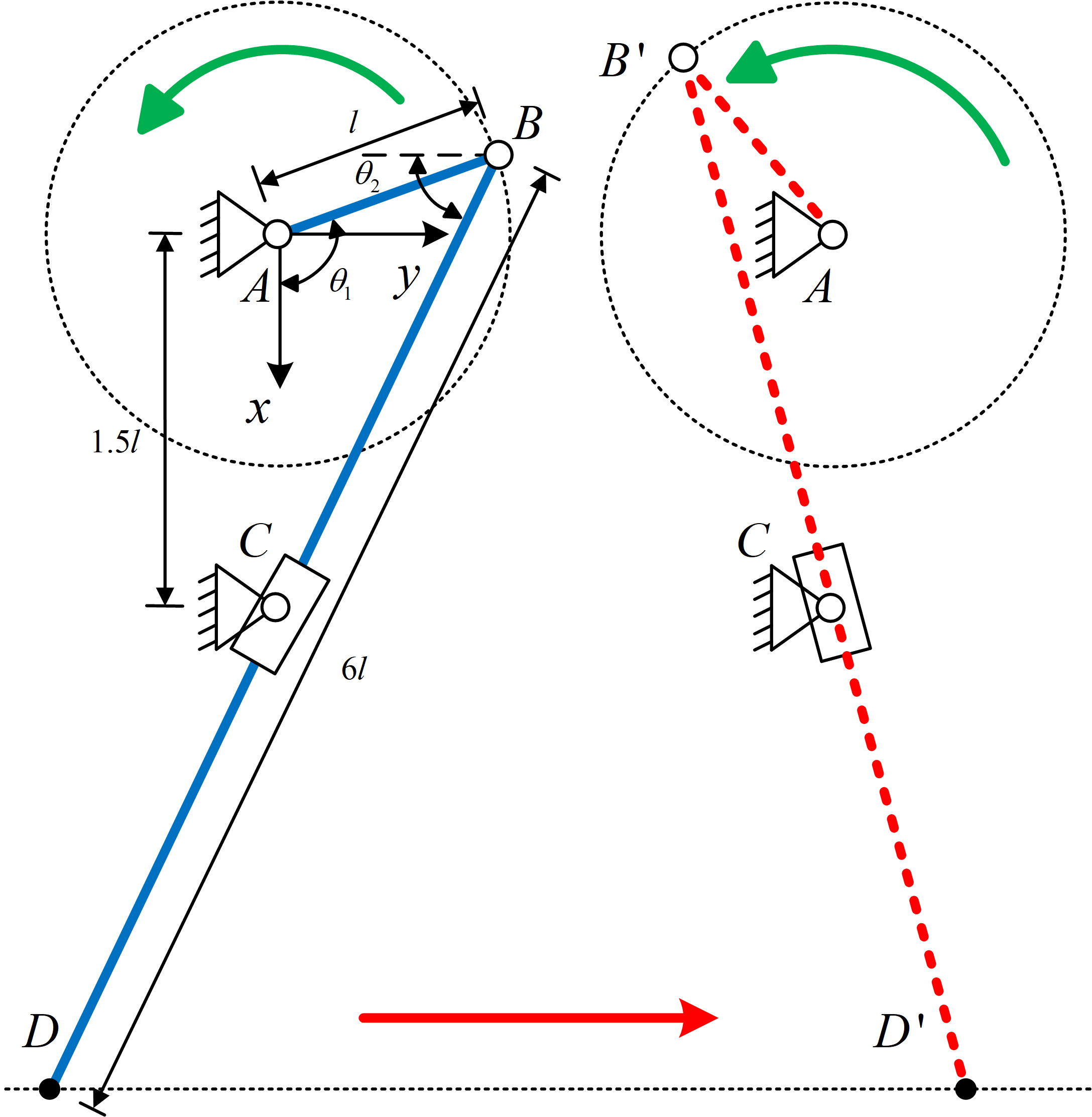}
    \caption {Motion Analysis of the Hoecken Linkage.}
    \label {fig:pathab}
\end {figure}

The kinematic model shown in Fig.~\ref{fig:pathab} defines the geometric parameters, where $l$ denotes the unit length with $l_{AB}=l$, $l_{AC}=1.5l$, and $l_{BD}=6l$. The rotation angles $\theta_1$ and $\theta_2$ correspond to the links $AB$ and $BD$, respectively. As illustrated in Fig.~\ref{fig:pathab}, the Hoecken link generates an approximately linear path of the fingertip during the rotation of $AB$ to $AB'$. The simulation results indicate that along this near-linear path, the maximum vertical deviation of point $D$ is 0.0164 units (about 0.492\,mm on the length of 220\,mm), which can be negligible for grasping tasks.

This near-linear motion ensures that the fingertip can travel horizontally in its default configuration, maintaining a constant height above the contact plane. The horizontal compliance effectively prevents unintended collisions with surfaces such as a tabletop during the initial approach, thereby improving grasp stability.

\begin{figure}[h]
    \centering
    \includegraphics[width=0.95\linewidth]{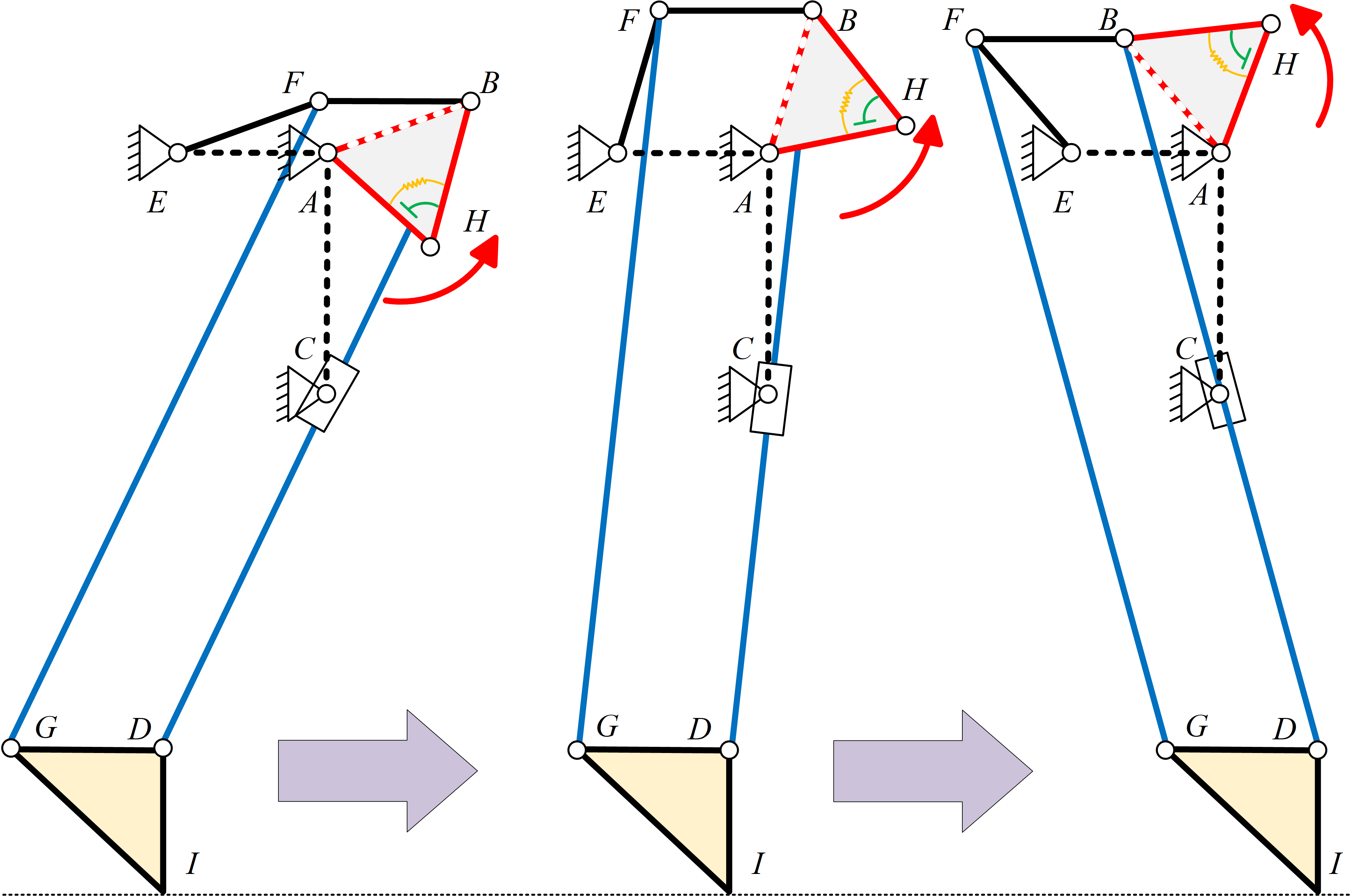}
    \caption{Horizontal parallel pinching motion.}
    \label{fig:horizon}
\end{figure}

As shown in Fig.~\ref{fig:horizon}, the link $HD$ has not yet contacted the object, and the distal phalanx $DI$ maintains a horizontal pinch.
Based on the Hoecken linkage (comprising links $AB$, $BD$, and slider at $C$), two parallelogram structures are introduced.

The first parallelogram $BDGF$ is formed by adding links $FG$ and $BF$, together with the finger segment $DG$. This arrangement transmits the angular relationship of $BF$ to $DG$, ensuring that when $BF$ is horizontal, $GD$ remains horizontal.

The second parallelogram $ABFE$ uses $AE$ as a fixed, horizontal reference on the base. A virtual link $AB$ is constrained via a tension spring and a mechanical stop between $AH$ and $BH$, maintaining a default angle $\angle AHB$ and keeping $AB$ parallel and equal in length to $EF$. Consequently, the horizontal reference from $AE$ is transmitted to $BF$, and subsequently to $GD$, allowing the distal phalanx to maintain a vertical orientation during horizontal pinching.

\subsection{Grasp Adaptation: Differential Linkage for Passive Enveloping Transition}

\begin{figure}[h]
    \centering
    \includegraphics[width=0.85\linewidth]{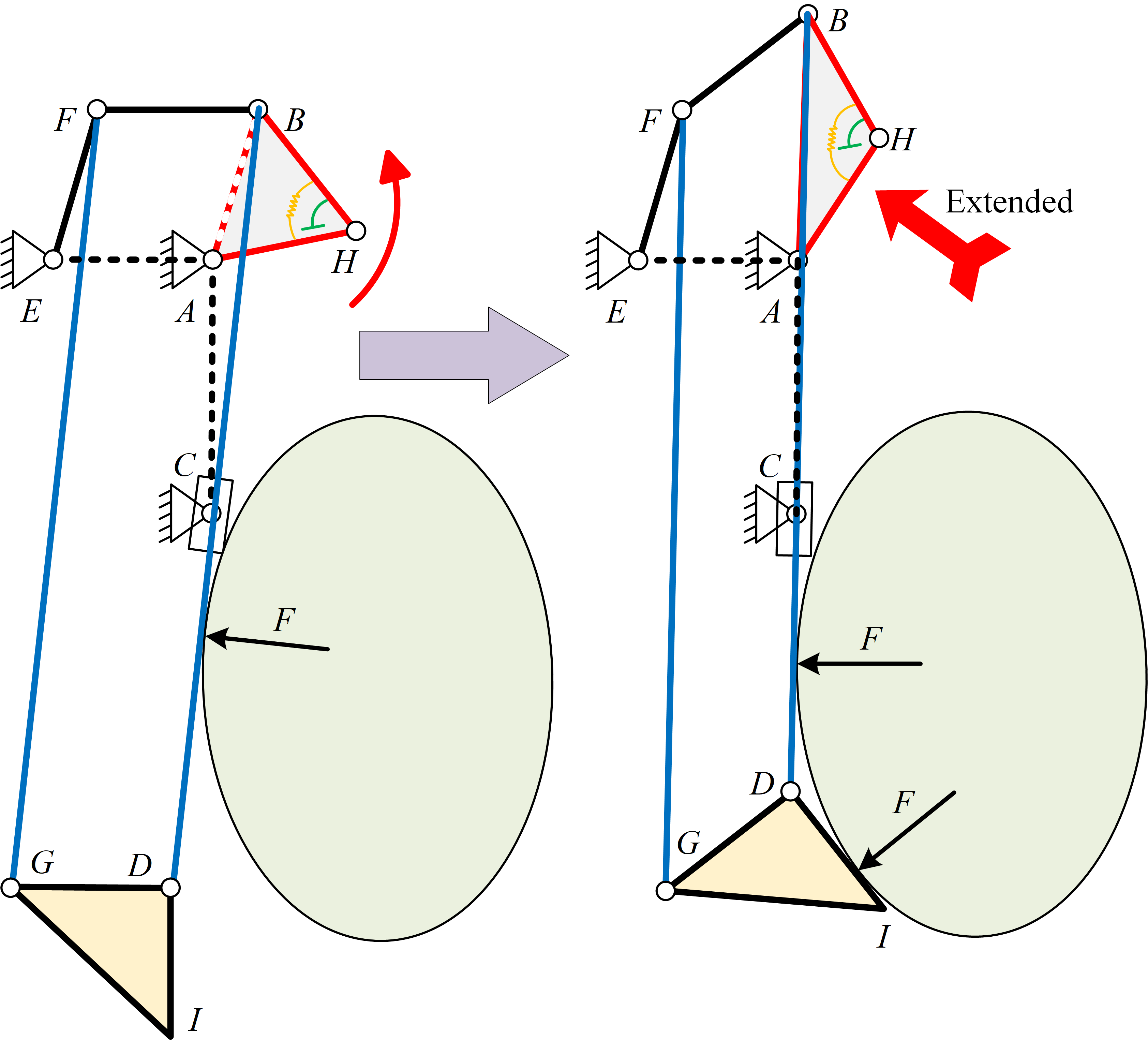}
    \caption{Enveloping grasp transition via differential linkage.}
    \label{fig:envelop}
\end{figure}

Building upon the horizontal parallel pinching motion shown in Fig.~\ref{fig:horizon}, the proposed design enables an automatic transition to an enveloping grasp when encountering large objects (Fig.~\ref{fig:envelop}). In the initial state, the distal phalanx performs parallel pinching under the constraints of the parallelogram mechanism, with $AH$ and $BH$ held together by a tension spring and limited by a stopper to maintain the parallelogram $ABFE$. 

Upon contact of $BD$ with the object, further horizontal motion is blocked, and the counterclockwise rotation of $A$ is no longer fully transmitted to $B$. Instead, due to the differential connection between $AH$ and $BH$, the two links are forced apart, altering the angle of $BF$, which is propagated through the linkage to $DG$, causing the distal phalanx to rotate inward and wrap around the object for a secure enveloping grasp.

When $BD$ makes contact with a large object and its motion is halted, continued rotation at pivot $A$ can no longer be transmitted directly along $BD$. Instead, the motion is diverted through $AH$ and $BH$ in the differential pair. This causes the angle between $AH$ and $BH$ to increase, stretching the spring and breaking the horizontal alignment of $BF$. This angular change is then transmitted through the linkage to $DG$, causing the distal phalanx to rotate inward and wrap around the object, passively achieving an enveloping grasp.

The corresponding dimensional parameters of the final design are provided in Table~\ref{tab:combined-params}.

\begin{table}[ht]
    \centering
    \caption{Dimensional parameters of the Hoecken-D Hand.}
    \label{tab:combined-params}
    \rowcolors{1}{blue!10}{yellow!20}
    \begin{tabular}{|c|c|c|c|c|c|}
        \hline
        Parameter & $l_{AC}$ & $l_{AH}$ & $l_{BH}$ & $l_{BD}$ & $l_{BF}$ \\ \hline
        Value & $30\,\text{mm}$ & $38\,\text{mm}$ & $38\,\text{mm}$ & $180\,\text{mm}$ & $30\,\text{mm}$ \\ \hline
    \end{tabular}
\end{table}

\subsection{3D Mechanism Embodiment: Parametric Modeling and Assembly Integration}

\begin{figure}[h]
    \centering
    \includegraphics[width=0.72\linewidth]{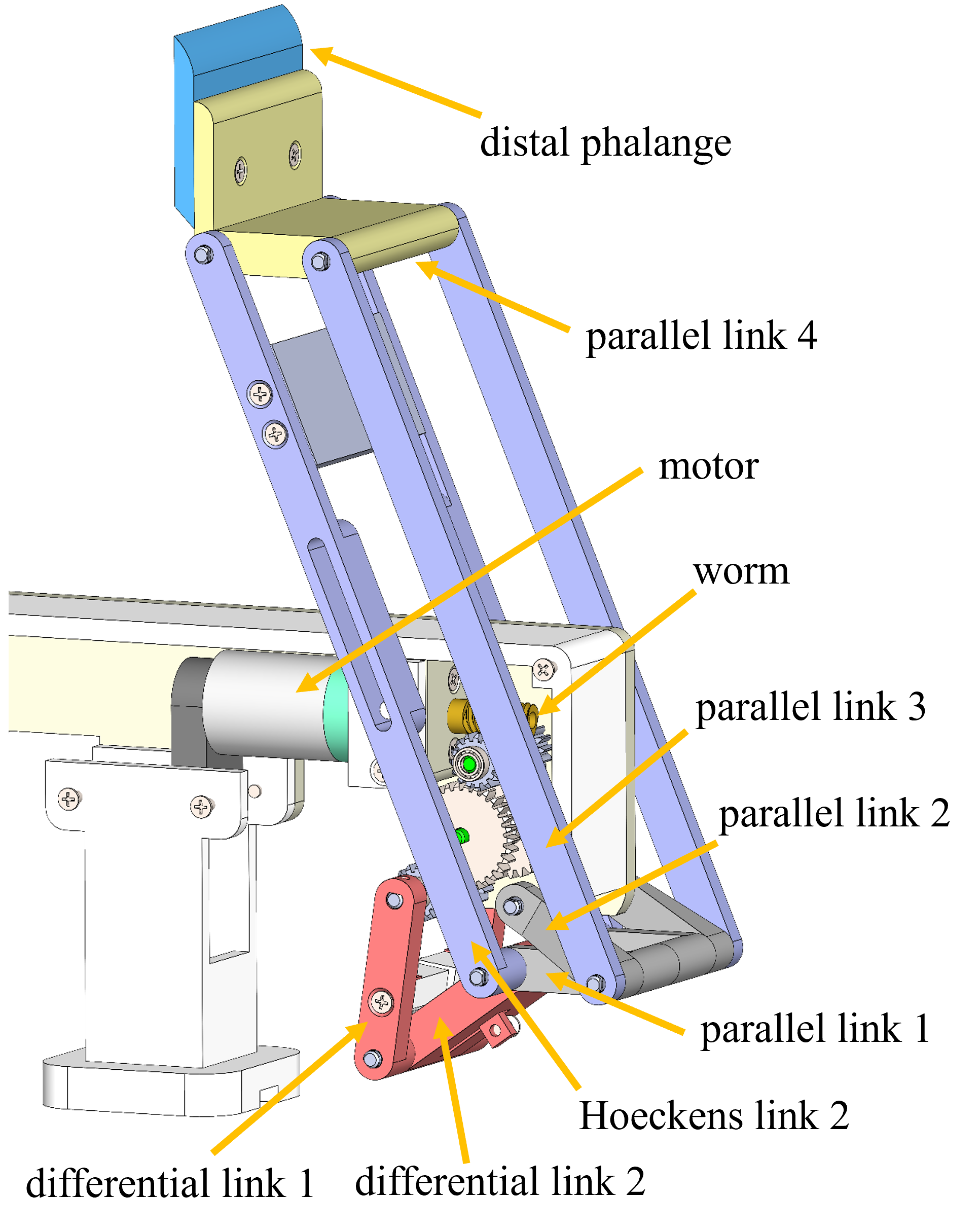}
    \caption{Configuration of the Hoecken-D Hand.}
    \label{fig:whole_hand}
\end{figure}

Through digital modeling, the complete mechanism of the Hoecken-D Hand was constructed through parametric modeling, ensuring accurate reproduction of the designed linkage relationships and motion constraints. The hand comprises two symmetrical fingers mounted on a common palm structure, with each finger integrating the Hoecken linkage, parallelogram, and differential mechanism. Fig.~\ref{fig:whole_hand} presents a detailed digital view of one finger.

The original Hoecken linkage consists of two members (Hoecken link~1 and Hoecken link~2) to generate an approximately vertical end-point trajectory. In our design, Hoecken link~1 is replaced by \emph{differential link~1} as the output rocker, paired with \emph{differential link~2} to preserve the straight-line guidance while introducing a differential spring system for compliance control.

Two parallelogram configurations are present.  

The first is a \emph{virtual parallelogram}: before Hoecken link~2 contacts the object, the two differential links remain closed, acting as a single Hoecken link~1. Together with the base, \emph{parallel link~1} and \emph{parallel link~2}, this transmits the base’s horizontal orientation to \emph{parallel link~2}. Once Hoecken link~2 makes contact, this parallelogram constraint is broken.  

The second is a \emph{physical parallelogram}: Hoecken link~2, \emph{parallel link~1}, \emph{parallel link~3}, and \emph{parallel link~4} form a closed loop that transfers the angle of \emph{parallel link~1} to \emph{parallel link~4}, driving the distal phalanx. During normal operation, the double parallelogram arrangement enables horizontal parallel pinching. When an object is encountered, Hoecken link~2 retracts, causing the distal phalanx to wrap inward for an enveloping grasp.

\begin{figure}[h]
    \centering
    \includegraphics[width=0.83\linewidth]{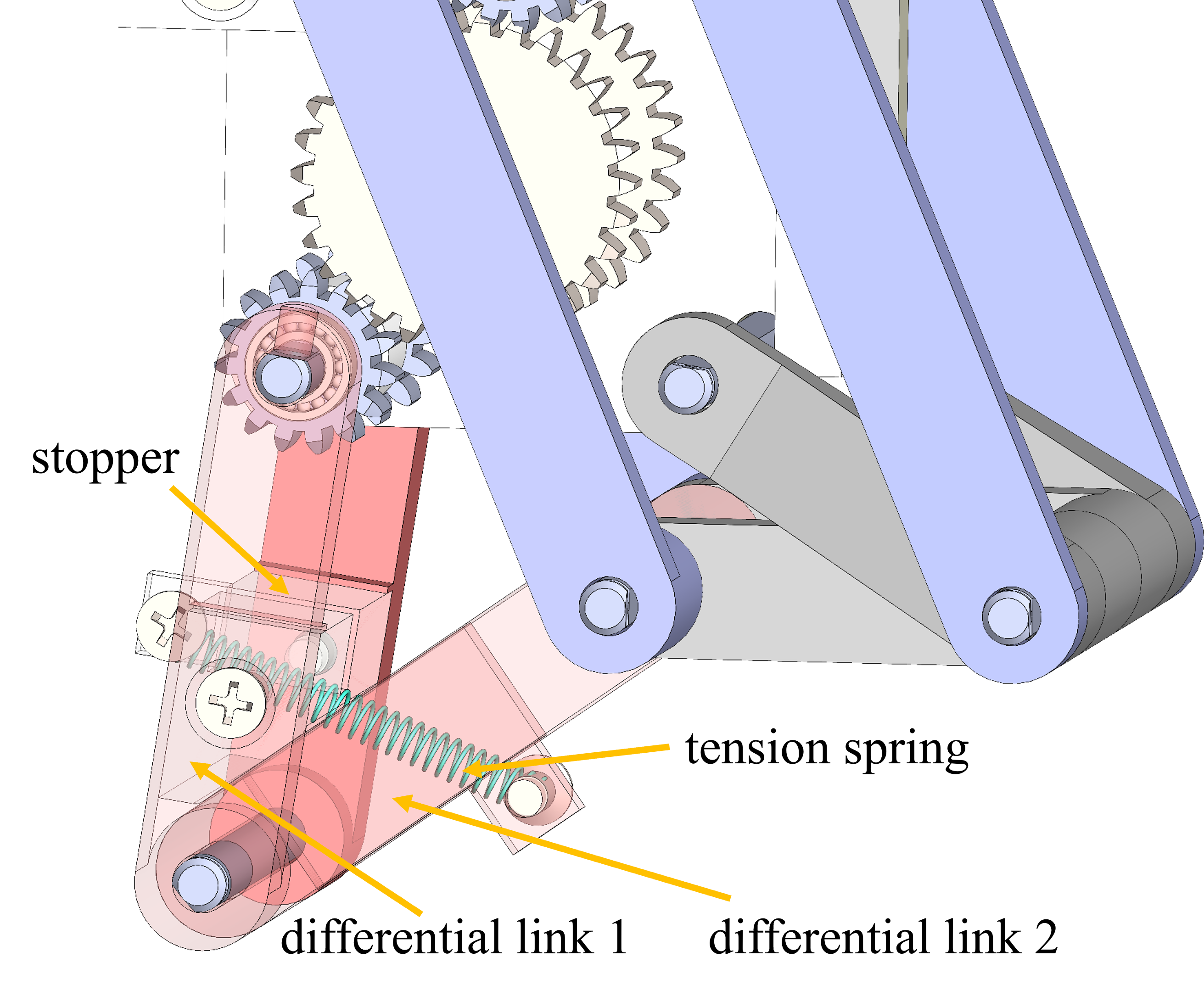}
    \caption{Detailed view of the differential mechanism.}
    \label{fig:differential}
\end{figure}

As shown in Fig.~\ref{fig:differential}, the two differential links (differential link~1 and differential link~2) are connected by a tension spring \(k\) and a stopper.  

Under normal operation, both links remain closed and act as a single member, transmitting motion to Hoecken link~2.  
When Hoecken link~2 is blocked by contact with an object, the output from the drive side is applied directly to differential link~1.  
This causes differential link~1 to continue rotating relative to differential link~2, stretching the tension spring until it overcomes the stopper constraint.  
At this point, the two links open, pulling Hoecken link~2 backward and initiating the inward wrapping motion of the distal phalanx for an enveloping grasp.

The stiffness and preload of the torsional spring are selected to avoid premature mode switching while ensuring reliable engagement once sufficient contact force is reached.

\section{ANALYSIS}

\subsection{Force Analysis in Linear Parallel Pinching}

During the linear parallel pinching phase of the Hoecken-D Hand, only the third phalanx (\(l_1 = 180~\mathrm{mm}\)) is in contact with the object.  
Unlike the simplified assumption in Watt-type fingers where the motor torque \(T_m\) is directly applied to the middle joint, our design introduces the Hoecken linkage between the actuator at point \(A\) and the finger linkage \(BD\).  
As a result, the effective torque at the middle joint is reduced or amplified depending on the instantaneous Jacobian of the linkage.

\begin{figure}[h]
    \centering
    \includegraphics[width=0.5\linewidth]{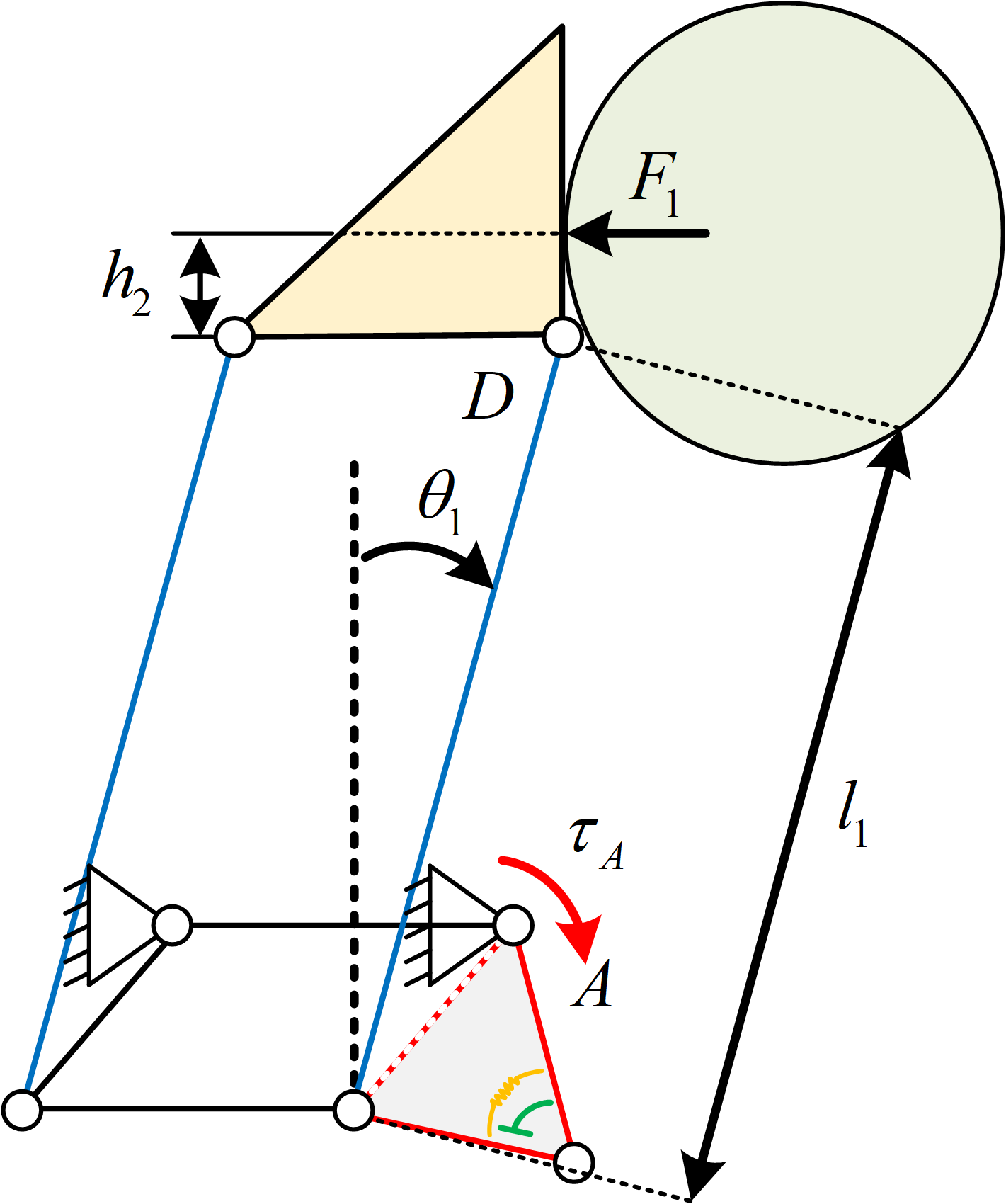}
    \caption{Force analysis on parallel grasping.}
    \label{fig:pinching}
\end{figure}

Let \(\tau_A\) be the motor output torque at point \(A\), and \(x_D(\theta_A)\) be the horizontal displacement of point \(D\) as a function of the rotation angle \(\theta_A\) of link \(AH\).  
From the principle of virtual work:
\begin{equation}
\tau_A \, \mathrm{d}\theta_A
=
F_{Dx} \, \mathrm{d}x_D
\quad\Rightarrow\quad
F_{Dx}(\theta_A) =
\frac{\tau_A}{\dfrac{\mathrm{d}x_D}{\mathrm{d}\theta_A}},
\label{eq:FDx}
\end{equation}
where \(F_{Dx}\) is the horizontal driving force at point \(D\).

Mapping \(F_{Dx}\) through the finger mechanism with an effective moment arm \(r_{\mathrm{eq}}(\theta)\) gives the middle joint torque:
\begin{equation}
M_{\mathrm{mid}}(\theta) = F_{Dx}(\theta_A) \cdot r_{\mathrm{eq}}(\theta).
\label{eq:Mmid}
\end{equation}

The grasping force \(F_1\) exerted by the third phalanx at a contact point located \(h_2\) away from the middle joint axis satisfies the moment balance:
\begin{equation}
\label{eq:F3_general}
F_1(h_2,\theta_1,\theta_A) =
\frac{M_{\mathrm{mid}}(\theta)}{h_2 + l_1\cos\theta_1}.
\end{equation}

For visualization purposes, if \(\frac{\mathrm{d}x_D}{\mathrm{d}\theta_A}\) and \(r_{\mathrm{eq}}\) are approximated as constants \(J_x\) and \(r_{\mathrm{eq}}\) respectively, Eq.~\eqref{eq:F3_general} simplifies to:
\begin{equation}
F_1(h_2,\theta_1) \approx
\frac{\tau_A \, r_{\mathrm{eq}}}{J_x \, \big(h_2 + 180\cos\theta_1\big)},
\label{eq:F3_simple}
\end{equation}
with \(h_2 \in [0, 50]~\mathrm{mm}\) and \(\theta_1 \in [0^\circ, 40^\circ]\).

\begin{figure}[h]
    \centering
    \includegraphics[width=0.8\linewidth]{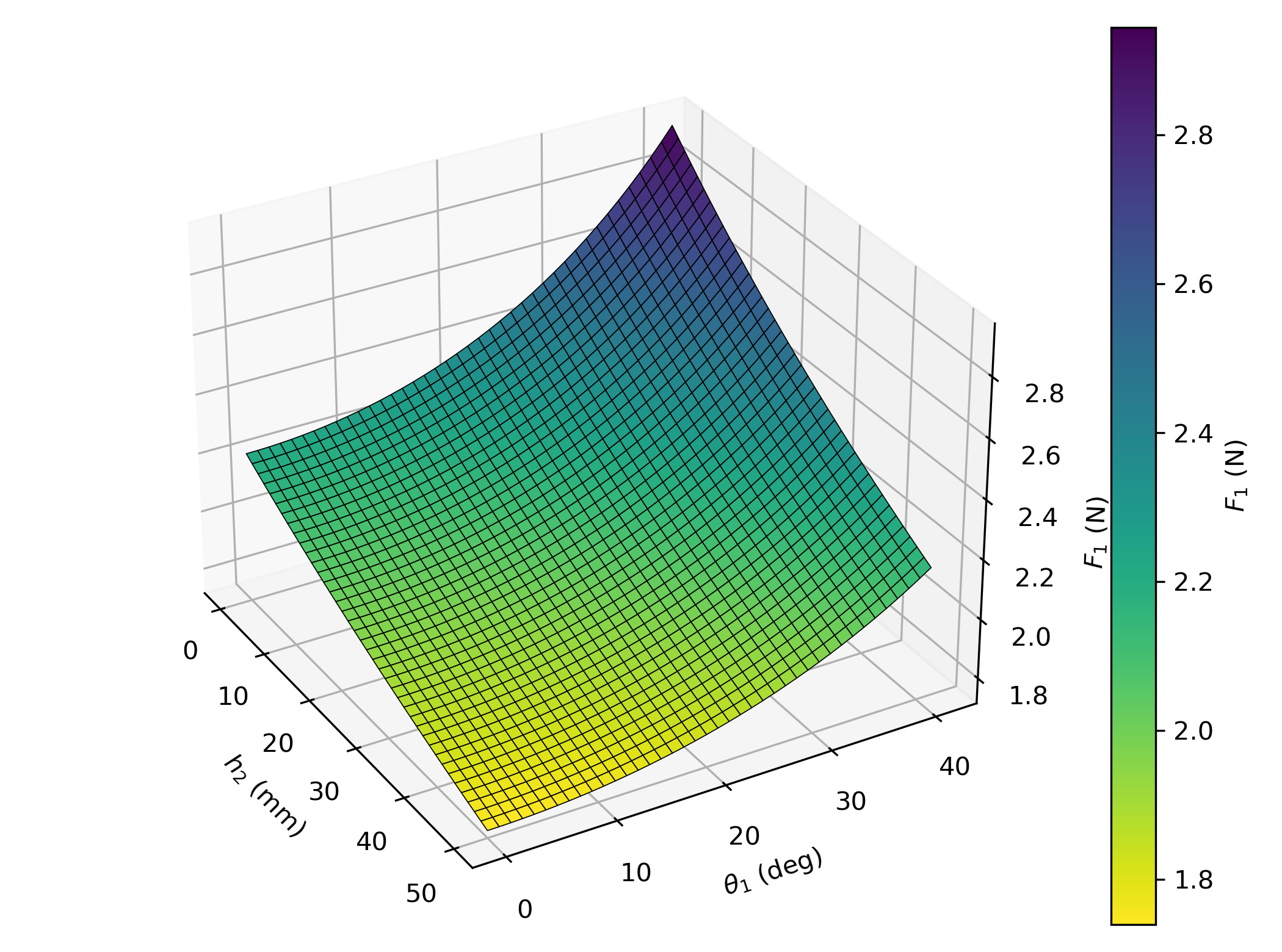}
    \caption{Grasping force $F_1$ vs.\ $\theta_1$ and $h_2$ during the linear parallel pinching.}
    \label{fig:pinching_surface}
\end{figure}

The surface in Fig.~\ref{fig:pinching_surface} confirms that $F_1$ grows with increasing $\theta_1$ (later in the stroke) and decreasing $h_2$ (contact point closer to the joint).  
This implies that for tasks requiring maximum pinch force, the hand should close until near the end of its travel and contact the object deeper within the phalanx rather than at the fingertip.

\subsection{Force Analysis in Self-Adaptive Enveloping}

\paragraph*{Generalized coordinates and contact geometry}
Let the generalized coordinates be 
$q=\begin{bmatrix}\theta_1 & \theta_2\end{bmatrix}^\top$, 
where $\theta_1,\theta_2$ are the rotations of the second and third phalanges relative to the vertical (positive inward). 
Denote normal contact forces by scalars $F_2,F_3$ and their Cartesian vectors by
\begin{equation}
\label{eq:F_vecs_new}
\mathbf{F}_2=
\begin{bmatrix}
F_2\cos\theta_1\\
-\,F_2\sin\theta_1
\end{bmatrix},
\qquad
\mathbf{F}_3=
\begin{bmatrix}
F_3\cos\theta_2\\
-\,F_3\sin\theta_2
\end{bmatrix}.
\end{equation}

With the middle joint as origin, the contact point positions are
\begin{equation}
\label{eq:G_points_new}
\mathbf{G}_1=
\begin{bmatrix}
h_1\sin\theta_1\\
h_1\cos\theta_1
\end{bmatrix},
\qquad
\mathbf{G}_2=
\begin{bmatrix}
l_1\sin\theta_1+h_2\sin\theta_2\\
l_1\cos\theta_1+h_2\cos\theta_2
\end{bmatrix},
\end{equation}
where $h_1,h_2$ are moment arms of the contact points to their joints and $l_1$ is the inter-joint lever for the third phalanx.

\begin{figure}[h]
    \centering
    \includegraphics[width=0.5\linewidth]{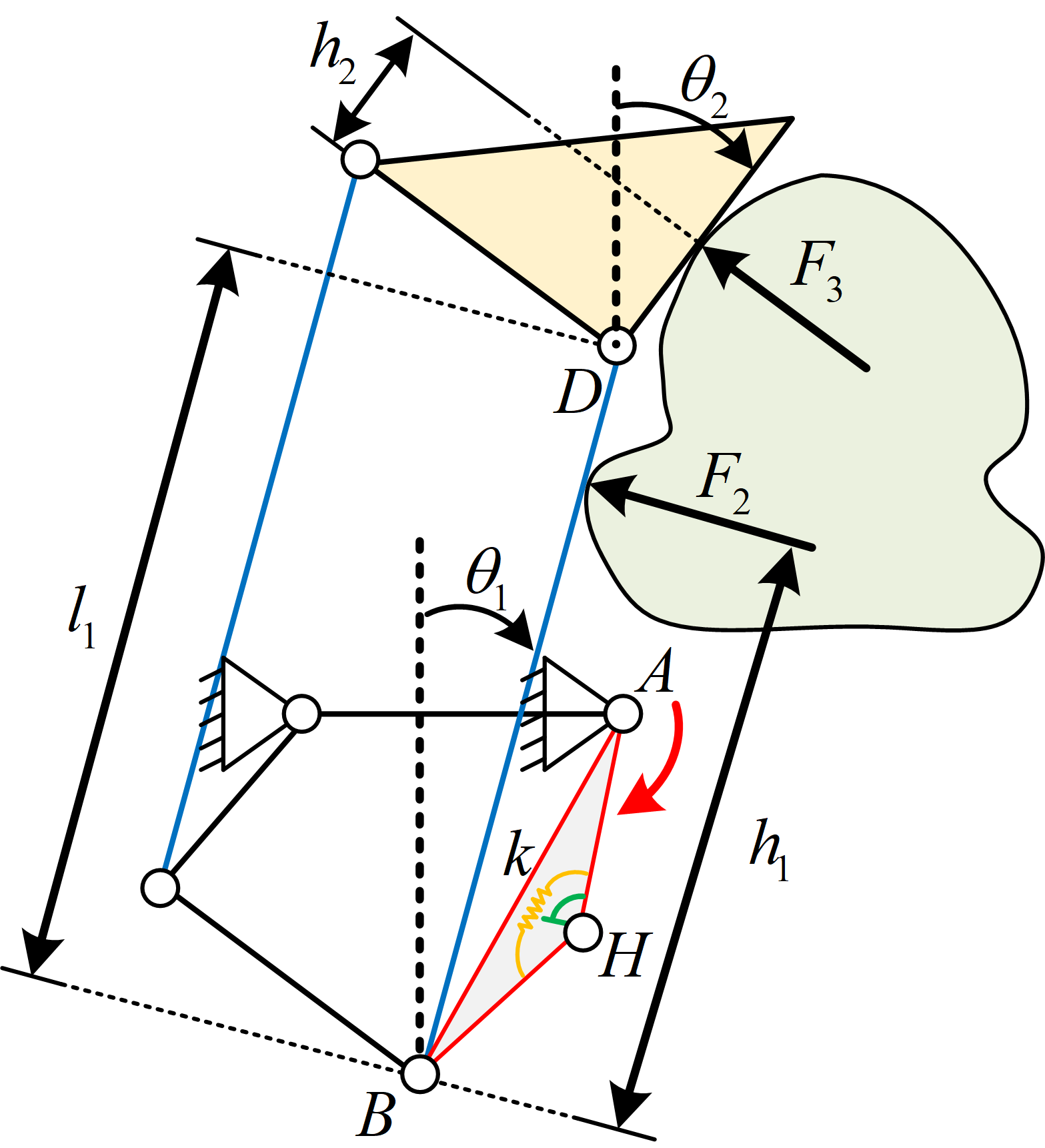}
    \caption{Force analysis on self-adaptive grasping.}
    \label{fig:adaptive}
\end{figure}

\paragraph*{Virtual displacements}
The virtual displacements under $\delta q$ are
\begin{align}
\label{eq:dG1_new}
\delta\mathbf{G}_1 &=
\begin{bmatrix}
h_1\cos\theta_1\\
-\,h_1\sin\theta_1
\end{bmatrix}\delta\theta_1,\\[3pt]
\label{eq:dG2_new}
\delta\mathbf{G}_2 &=
\begin{bmatrix}
l_1\cos\theta_1\\
-\,l_1\sin\theta_1
\end{bmatrix}\delta\theta_1
+
\begin{bmatrix}
h_2\cos\theta_2\\
-\,h_2\sin\theta_2
\end{bmatrix}\delta\theta_2.
\end{align}

\paragraph*{Differential torsion from geometry}
Let $A=(0,0)$, $C=(0,45)\,\mathrm{mm}$, and $E=(-30,0)\,\mathrm{mm}$. 
The straight-line stage constrains $EF=30\,\mathrm{mm}$, parallel to the initial $AB$, and locks $F$ after first contact. 
For a given $\theta_1$, the pre-contact point $B_0$ and $F$ are determined from the geometric constraints. 
When an enveloping rotation $\theta_2$ is applied, point $B$ moves along $DB$ such that $BF$ makes angle $\theta_2$ with the $x$-axis. 
The virtual bar $AB$ has length $\|B(\theta_1,\theta_2)-A\|$.

The differential pair $AH=BH=38\,\mathrm{mm}$ is connected by a torsional spring at hinge angle
\begin{equation}
\label{eq:alpha_simple}
\alpha(\theta_1,\theta_2) 
= \arccos\!\left(\frac{AH^2 + BH^2 - AB^2}{2\,AH\,BH}\right).
\end{equation}
Relative to its pre-contact angle $\alpha_0(\theta_1)$, the spring torque is
\begin{equation}
\label{eq:tau_d_simple}
\tau_d(\theta_1,\theta_2) = k_d\,[\alpha(\theta_1,\theta_2) - \alpha_0(\theta_1)].
\end{equation}

\paragraph*{Force balance via virtual work}
The generalized spring torques are $(-\tau_d s_1,\ -\tau_d s_2)$, with $s_i = \partial\alpha/\partial\theta_i$. 
The total virtual work vanishes:
\begin{equation}
\label{eq:virt_work_new}
\underbrace{\big(\tau_1-\tau_d s_1\big)\delta\theta_1 + \big(-\tau_d s_2\big)\delta\theta_2}_{\text{actuation \& spring}}
=
\underbrace{\mathbf{F}_2^\top\delta\mathbf{G}_1 + \mathbf{F}_3^\top\delta\mathbf{G}_2}_{\text{contact work}}.
\end{equation}
Using the kinematic Jacobians for the contact points, this becomes
\begin{equation}
\label{eq:balance_matrix_simple}
\begin{bmatrix}
h_1 & l_1\cos(\theta_2-\theta_1)\\
0   & h_2
\end{bmatrix}
\begin{bmatrix}
F_2\\ F_3
\end{bmatrix}
=
\begin{bmatrix}
\tau_1 - \tau_d s_1\\
-\tau_d s_2
\end{bmatrix}.
\end{equation}

\paragraph*{Closed-form forces}
From \eqref{eq:balance_matrix_simple}, the forces are
\begin{align}
\label{eq:F3_closed_simple}
F_3(\theta_1,\theta_2) &= -\,\frac{\tau_d(\theta_1,\theta_2)\,s_2(\theta_1,\theta_2)}{h_2}, \\[4pt]
\label{eq:F2_closed_simple}
F_2(\theta_1,\theta_2) &= \frac{\tau_1 - \tau_d(\theta_1,\theta_2) \, s_1(\theta_1,\theta_2)}{h_1} \notag \\
&\quad - \frac{l_1\cos(\theta_2-\theta_1)}{h_1} \, F_3(\theta_1,\theta_2).
\end{align}
Both forces increase with actuation torque $\tau_1$ and with the spring opening $\alpha-\alpha_0$, modulated by the geometric Jacobians $s_1,s_2$.

Angles are in radians in all formulas, with plots converted to degrees.  
The functions $\alpha(\theta_1,\theta_2)$, $s_1(\theta_1,\theta_2)$, and $s_2(\theta_1,\theta_2)$ are obtained from \eqref{eq:alpha_simple}—\eqref{eq:tau_d_simple} and \eqref{eq:balance_matrix_simple}—\eqref{eq:F2_closed_simple}.

\begin{figure}[h]
    \centering
    \includegraphics[width=0.80\linewidth]{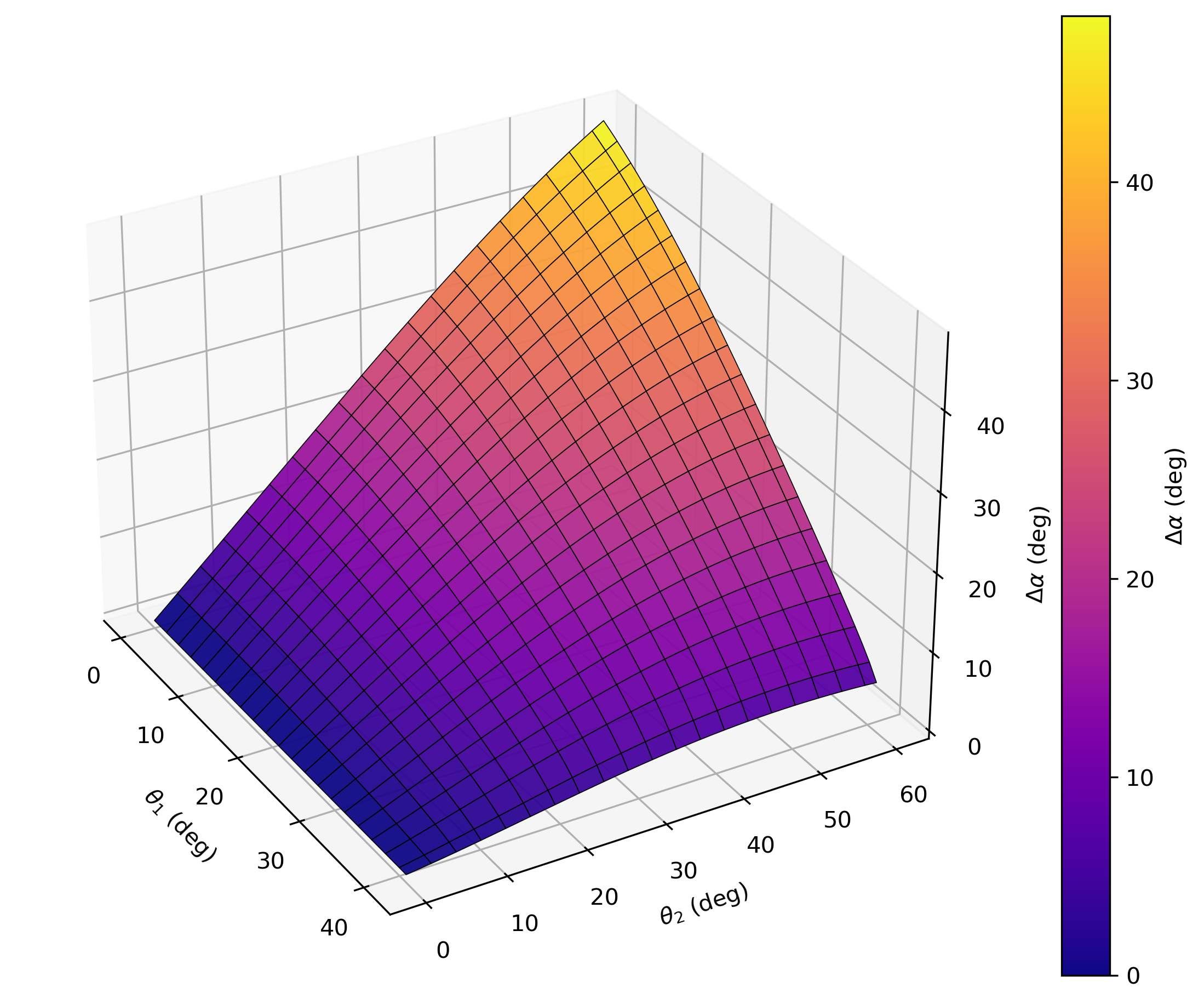}
    \caption{Spring opening angle $\alpha$ vs.\ $\theta_1$ and $\theta_2$.}
    \label{fig:spring_opening}
\end{figure}

Figure~\ref{fig:spring_opening} shows the spring opening angle $\alpha$ over the feasible $(\theta_1,\theta_2)$ range for the nominal parameters in Table~\ref{tab:parameters}.  
Larger $\alpha$ means greater spring preload and thus higher restoring torque.  
Increasing either $\theta_1$ or $\theta_2$ generally raises $\alpha$, but the relationship is nonlinear, with stronger coupling at higher $\theta_1$.

\begin{figure}[h]
    \centering
    \includegraphics[width=0.80\linewidth]{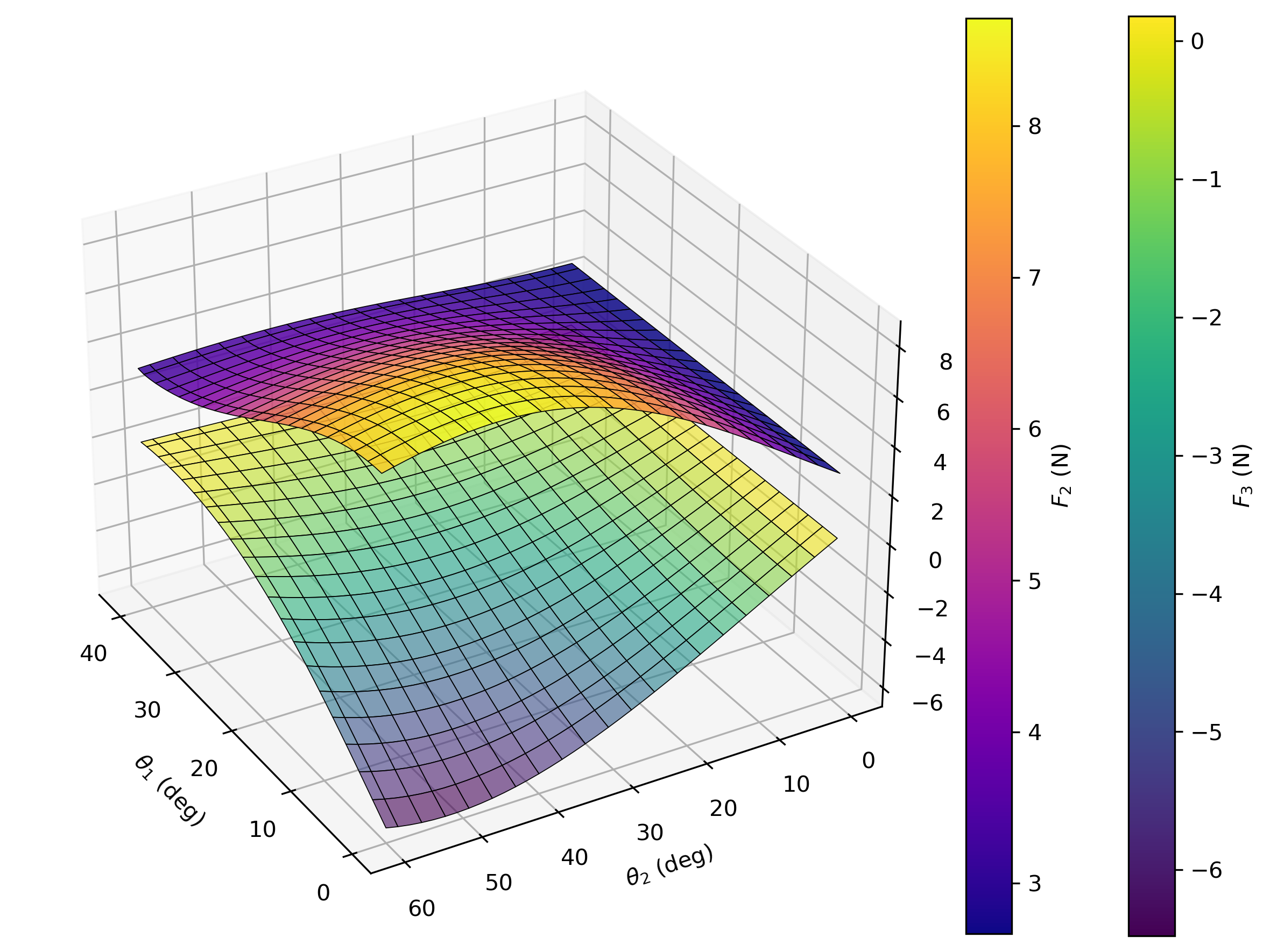}
    \caption{Self-adaptive enveloping forces $F_2$ and $F_3$ vs.\ $\theta_1$ and $\theta_2$.}
    \label{fig:enveloping_forces}
\end{figure}

Figure~\ref{fig:enveloping_forces} shows the contact force components $F_2$ and $F_3$ computed from \eqref{eq:F3_closed_simple}—\eqref{eq:F2_closed_simple} using the spring deflection in Fig.~\ref{fig:spring_opening}.  
The plots reveal a load redistribution between the two contact directions: $F_2$ grows with $\theta_2$ while $F_3$ becomes more negative, reflecting a trade-off in force capacity with grasp configuration.

\paragraph*{Parameterization for plots}
The parameters used in the plots are listed in Table~\ref{tab:parameters}.

\begin{table}[hpb]
    \centering
    \caption{Geometric and mechanical parameters used in simulations.}
    \label{tab:parameters}
    \rowcolors{1}{blue!10}{yellow!20}
    \begin{tabular}{|c|c|l|}
        \hline
        Parameter & Value & Description \\ \hline
        $l_1$     & $180~\mathrm{mm}$ & BD length \\
        $h_1$     & $150~\mathrm{mm}$ & Initial vertical offset of D \\
        $h_2$     & $40~\mathrm{mm}$  & Offset from C to B along BD \\
        $AH=BH$   & $38~\mathrm{mm}$  & Differential arm lengths \\
        $k_d$     & $800~\mathrm{N\cdot mm/rad}$ & Torsional spring stiffness \\
        $\tau_1$  & $400~\mathrm{N\cdot mm}$     & Applied torque at joint 1 \\
        $AB_0$    & $30~\mathrm{mm}$             & Initial AB length \\ \hline
    \end{tabular}
\end{table}

\section{Experiments}

To verify the feasibility and adaptability of the Hoecken-D Hand, a full-scale prototype was fabricated via PLA-based 3D printing. The assembled hand preserves the designed linkage geometry and spring arrangements, achieving a measured parallel pinching span of approximately 0—200~mm. We evaluated its performance in two representative grasping modes:  
(1) \textbf{Pure parallel pinching}—both fingers maintain the double-parallelogram constraint, keeping the fingertips parallel throughout closure; and
(2) \textbf{Mid-stroke transition to enveloping}—one finger’s Hoecken link~2 is blocked by the object, triggering the differential spring to rotate the fingertip inward for adaptive wrapping.

\begin{figure}[h]
    \centering
    \includegraphics[width=0.95\linewidth]{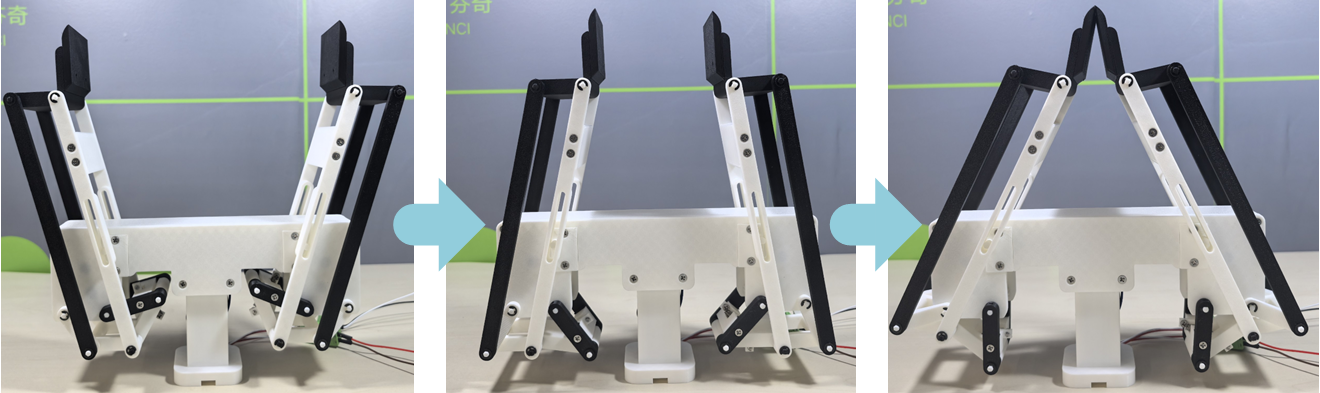}
    \caption{Sequence showing the Hoecken-D Hand transitioning from pure parallel pinching (left) to adaptive enveloping (right) as one finger’s motion is blocked and the differential spring mechanism engages.}
    \label{fig:transition}
\end{figure}

In tests with a diverse set of objects—including thin sheets (thickness 0.5—2~mm), bottles (diameter 50—80~mm), mugs, boxes (width 40—110~mm), and soft packages—the hand achieved a $>95\%$ success rate for rigid objects within the parallel-pinching range, with most failures occurring for very thin sheets due to alignment errors. The mid-stroke transition strategy significantly improved handling of such thin items, raising the success rate to over $88\%$ compared to below $50\%$ with pure parallel pinching. For objects between 60—100~mm in diameter, the enveloping motion achieved $90\%$ success; larger objects were limited by the maximum opening span. These results confirm that the differential spring mechanism effectively extends the hand’s adaptability without compromising basic parallel pinching stability.

\begin{figure}[h]
    \centering
    \includegraphics[width=0.95\linewidth]{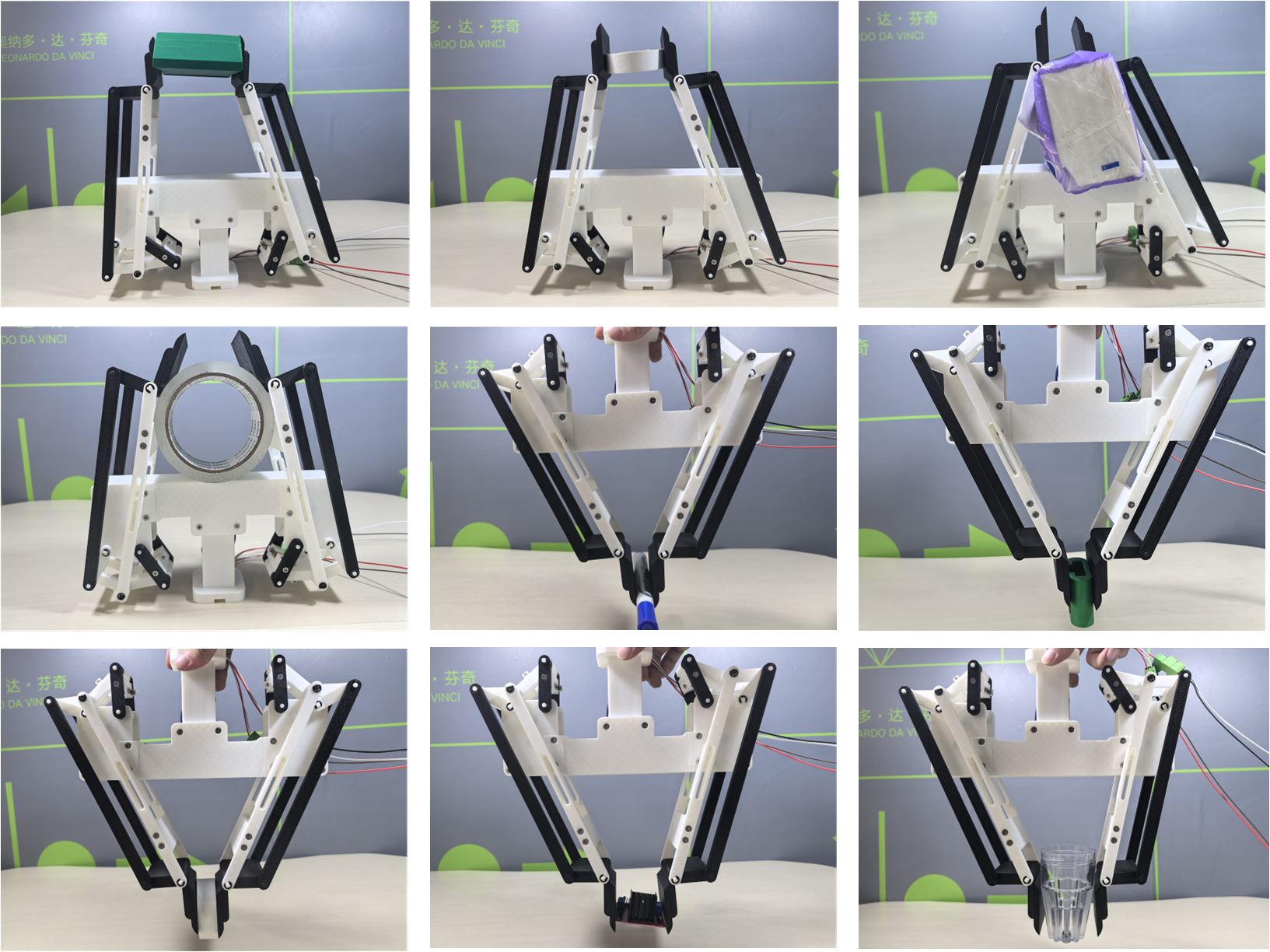}
    \caption{Representative grasps with the Hoecken-D Hand, including thin sheets, cylindrical and box-like objects, soft packages, and irregular shapes, across both parallel pinching and mid-stroke enveloping modes.}
    \label{fig:demo_collage}
\end{figure}

A broader range of trials is shown in Fig.~\ref{fig:demo_collage}, demonstrating stable grasps across various shapes, sizes, and surface properties. The experiments demonstrate that the Hoecken-D Hand can reliably switch between precise linear pinching and adaptive enveloping, matching the grasping versatility predicted by the kinematic and force analyses.

\section{Conclusions}
We presented the Hoecken-D Hand, an underactuated gripper that combines a modified Hoecken linkage with a differential spring-stopper pair to realize two complementary modes: precise linear parallel pinching and a passive, contact-triggered transition to adaptive enveloping. The double-parallelogram arrangement preserves fingertip parallelism during pinching, while the differential pair reorients the distal phalanx upon obstruction, improving stability on irregular and thin objects—without additional sensors or actuators. The mechanism admits a single-linear-actuator drive for two fingers (our prototype uses one actuator per finger for simplicity), and our kinematic and force analyses clarify how geometry and spring deflection shape grasp forces and switching behavior. Prototype results across diverse objects corroborate the design’s versatility, highlighting a compact, low-cost, and mechanically intelligent solution for unstructured environments. Looking ahead, we will (i) integrate a single-actuator synchronization scheme, (ii) formalize stiffness/preload selection and tolerance-to-behavior sensitivity, (iii) add quantitative benchmarking of grasp forces, switching thresholds, and repeatability, and (iv) assess durability under extended use.

\addtolength{\textheight}{-12cm}   


\begin{thebibliography}{99}

\bibitem{MIThand} S. Jacobsen, E. Iversen, D. Knutti \textit{et al.}, “Design of the Utah/M.I.T. Dextrous Hand,” \textit{IEEE Int. Conf. on Robotics and Automation}, San Francisco, USA, April, 1986.

\bibitem{DLRhand2008} H. Liu \textit{et al.}, “Multisensory five-finger dexterous hand: The DLR/HIT Hand II,” \textit{IEEE/RSJ Int. Conf. on Intelligent Robots and Systems}, Nice, France, March, 2008.

\bibitem{JPLhand1987} C. Loucks, V. Johnson, P. Boissiere \textit{et al.},  “Modeling and Control of the Stanford/JPL Hand,” \textit{IEEE Int. Conf. on Robotics and Automation}, Raleigh, USA, April, 1987.

\bibitem{Robonauthand} C. S. Lovchik, M. A. Diftler, “The Robonaut hand: a Dexterous Robot Hand for Space,”  \textit{IEEE Int. Conf. on Robotics and Automation }, Detroit, USA, April, 1999.

\bibitem{PASA2016} Liang, D., Song, J., Zhang, W., \textit{et al}. “PASA Hand: A Novel Parallel and Self-Adaptive Underactuated Hand with Gear-Link Mechanisms”. \textit{Int. Conf. on Intelligent Robotics and Applications.} vol 9834., 2016.

\bibitem{SARAH} Martin E., Desbiens A. L., Lalibert E, T., \textit{et al.}, “SARAH Hand Used for Space Operations on STVF Robot,” \textit{Int. Conf. on Intelligent Manipulation and Grasping}, Genoa, Italy, July 1-2, 2004: 325-331.

\bibitem{Barrett} Townsend W. “The Barrett Hand Grasper-Programmably Flexible Part Handling and Assembly.” \textit{Industrial Robot: An Int. J.}, 2000, 27(3), pp. 181-188.

\bibitem{robotiq} Demers A., Lefrançois L. A., Jobin S., \textit{et al.}, “Gripper Having a Two Degree of Freedom Underactuated Mechanical Finger for Encompassing and Pinch Grasping”. US Patent. US8973958B2. 2015-03-10.

\bibitem{SDM} A. M. Dollar and R. D. Howe, “The SDM Hand as a Prosthetic Terminal Device: A Feasibility Study,” \textit{IEEE Int. Conf. on Rehabilitation Robotics}, Noordwijk, Netherlands, 2007.

\bibitem{chebyshev2017} J. Xu \textit{et al.}, “LPSA Underactuated Mode of Linearly Parallel and Self-adaptive Grasping in the CLIS Robot Hand with Chebyshev Linkage and Idle Stroke,” \textit{Int. Conf. on Advanced Robotics and Mechatronics (ICARM)}, Hefei and Tai'an, China, 2017.

\bibitem{Liu2019} Y. Liu and W. Zhang, “An Underactuated Robot Finger with Movable Skeleton Link and Hoecken Mechanism for Linear Clamping and Self-adaptive Grasping,” \textit{IEEE Int. Conf. on Real-time Computing and Robotics (RCAR)}, Irkutsk, Russia, 2019.

\bibitem{Feng2024} K. Feng, Z. Duan, C. Han, \textit{et al.}, “A Novel Robot Finger Geometry for Parallel Self-Adaptive Grasp with Rack-Crank-Slider Mechanism,” \textit{IEEE Int. Conf. on Robotics, Automation and Mechatronics}, Hangzhou, China, 2024.

\bibitem{Kim2022} D. Yoon and K. Kim, “Fully Passive Robotic Finger for Human-Inspired Adaptive Grasping in Environmental Constraints,” \textit{IEEE/ASME Trans. on Mechatronics}, 27(5), Oct. 2022.

\bibitem{Liang2023} B. Liang, Y. Tian , W. Zhang , “A Straight Parallel Pinch and Self-adaptive Gripper with Differential and Watt Linkages,” \textit{ROBOT}, 45(6), Nov. 2023.

\bibitem{Liu2023} Y. Liu, W. Zhang , “A Robot Gripper with Differential and Hoecken Linkages for Straight Parallel Pinch and Self-Adaptive Grasp,” \textit{Appl. Sci.}, 13(12), 2023.


\end{thebibliography}
\end{document}